\documentclass[11pt,a4paper,logo]{lumia}

\usepackage[numbers,sort&compress]{natbib}
\usepackage{etoolbox}
\usepackage{nicefrac}
\usepackage{subcaption}

\hypersetup{colorlinks=true, linkcolor=blue!55!black, citecolor=teal!60!black, urlcolor=blue!55!black}

\definecolor{abstrabg}{HTML}{F2D1C5}
\definecolor{black50}{gray}{0.5}

\definecolor{citecolor}{HTML}{0071bc}

\hypersetup{
    colorlinks=true,
    linkcolor=red,
    citecolor=citecolor,
    filecolor=magenta,      
    urlcolor=magenta,
}
\newcommand{\Qstar}{Q^{\!*}}

\newcommand{\Rf}{R_{\mathrm{f}}}
\providecommand{\RLHEV}{RLHEV}

\title{Agentic Game Development as a Verifiable Trajectory Data Engine for Scaling World Models}

\setheadertitle{Agentic Game Development as a Verifiable Trajectory Data Engine for Scaling World Models}

\author{%
    \parbox{0.75\textwidth}{\centering
    Pengfei Zhou$^{1,2}$ \quad
    Hexin Wang$^{2}$ \quad
    Zhengfeiyang Zhang$^{2}$ \quad
    Yixing Ma$^{3}$ \quad
    Zhenglin Wan$^{1}$ \quad 
    Kaipeng Zhang$^{4}$ \quad
    Wangbo Zhao$^{5}$$^\ddagger$ \quad
    Yang You$^{1}$$^\ddagger$ \\
    }\\
    \vspace{-0.5em}
     $^{1}$National University of Singapore, HPC-AI Lab \quad 
     $^{2}$InfRec, Cardinal AI Lab \quad
     $^{3}$University of California, Berkeley 
     $^{4}$Independent Researcher \quad 
     $^{5}$Hong Kong University of Science and Technology \\ 
}

\correspondingemail{\emailicon~\href{mailto:wangbo.zhao96@gmail.com}{wangbo.zhao96@gmail.com}; \emailicon~\href{mailto:yangyou@nus.edu.sg}{yangyou@nus.edu.sg} \quad $^\ddagger$ Corresponding Author
}

\begin{document}

\begin{abstract}
A common strategy for scaling world models is to train on more crawled video with more compute. We argue that this strategy is inefficient: scaling world models also requires a recursive data engine that offers grounded reward signals. The success of code agents illustrates why this matters. As code is executable, compilers and runtimes can provide high-quality rewards for Reinforcement Learning (RL) post-training of LLMs. By contrast, spatial generation still relies largely on fuzzy proxies such as CLIP scores. These signals are fuzzy and biased, making them hard to support RL post-training. Compared with these, game development provides a missing reward environment for spatial world models. A scene encoded by a game engine is an executable world specification: the engine can efficiently check collision, physics, navigability and bounded playability, while the developer provides the global verification signal by judging whether the scene should be accepted. Game development also provides real-world long-horizon trajectory data for RL post-training. We therefore propose Reinforcement Learning with Human-Engine Verification (RLHEV), a post-training paradigm that combines dense engine signals with implicit human acceptance feedback from the development process. We apply this training objective to our proposed Agentic World Model (AWoMo): a world-building agent that proposes scene edits, observes human-engine verification, and converts accepted or repaired multimodal traces into training data. We evaluate the proposed approach through controlled experiments. On UnitySceneBench, a 200-example Unity asset-edit evaluation, our RLHEV obtains the highest score. In generalization, transfer learning helps with out-of-distribution shifts and gives positive signals in Unreal and Godot cross-engine experiments. AWoMo-augmented training also improves the embodied performance of the policy on R2R, Gymnasium MuJoCo, and D4RL Gym-MuJoCo. Agentic artifacts are released for reproduction: \href{https://github.com/LanceZPF/cardinal-preview}{https://github.com/LanceZPF/cardinal-preview}.
\end{abstract}

\maketitle

\section{Introduction}
\label{sec:intro}

Recent world-model work often treats spatial intelligence as a scaling problem: collect more scraped video, train larger models, and spend more compute~\citep{brooks2024sora,bruce2024genie}. Code agents illustrate why this recipe is inefficient. Beyond foundation models pretrained on large-scale corpora, code agents have become one of the most important LLM applications today~\citep{yang2024sweagent}. \textit{Why are code agents winning the game?}

The key is that code is executable and functional, combining two signals that spatial generation usually separates. 
First, code can be executed: compilers, tests, and runtimes provide dense, low-cost feedback~\citep{li2022alphacode}.
Second, the developer provides supplementary verification: a patch that passes tests can still be rejected because it breaks maintainability or fails the product goal in an agentic workflow. The post-training advantage of coding agents comes from this human-compiler dual verification. Such verification feedback enables reinforcement learning to continuously improve the model after pretraining, incentivizing not only coding ability but also general reasoning and planning~\citep{deepseek2025r1}.

Spatial intelligence is on the other side. The quality of multimodal spatial outputs, such as generated videos or 3D scenes, is usually scored by fuzzy proxies such as JSD~\citep{menendez1997jensen}, FVD~\citep{unterthiner2018fvd}, CLIP similarity~\citep{radford2021clip,ljungbergh2026r3d2,wu2026unilat3d}, and MLLM-as-judge scores~\citep{chen2024mllm,xia2026sage,lu20264dworldbench}. These proxies are noisy, biased, and gameable~\citep{gao2023overoptimization,skalse2022rewardhacking,chengposition2025}. Human judgment is indispensable, but if it is collected only as subjective preference labels over final outputs, it is too expensive and low-bandwidth to support the kind of iterative post-training loop that made code and reasoning scale.

Our central argument is that spatial intelligence is hitting a data-centric version of the Bitter Lesson~\cite{sutton2019bitter,ying2026beyond}. Scraping more data can improve coverage, but it does not create a reward engine that automatically provides high-quality supervision. 
We argue that game development provides such a recursive data engine: it records how humans build worlds, captures engine- and human-generated verification signals throughout the process, and converts these signals into post-training data. 

Thus, we propose \emph{Agentic World Model} (AWoMo) for automating game development: a world model embedded in a developer-centered agent workflow that proposes, renders, checks, revises, and learns from world-building traces~\citep{chu2026agenticworldmodeling}. AWoMo facilitates storing each prompt, scene program, rendered state, failure, fix, engine check, and human decision as a complete multimodal world-building trajectory. These trajectories enable recursive evolution: better models build more candidate worlds, human-engine verification supplies reward, and accumulated traces train the next model.

\begin{quote}
\textbf{Game development matters for scaling world models since it can turn world construction into a recursive feedback engine.}
Specifically, world models need a loop that observes how worlds are built, checks whether intermediate results work, and turns developer decisions into supervision, whereas game development offers such a loop. Engines can automatically test whether generated worlds are structurally valid, while developers judge whether they are truly qualified. This provides a dual verification signal: engine checks provide dense structural rewards, while developer feedback provides sparse but highly aligned judgments on acceptance.
Based on this, we propose AWoMo and train it with RLHEV: Reinforcement Learning with Human-Engine Verification.
Under RLHEV, game development trajectories give AWoMo a path similar to coding models: execution exposes local mistakes, humans judge global success, those signals improve the model, and improved models generate better candidates. This creates a possibility of a self-improving loop for world models, while unifying spatial understanding and generation capabilities.
\end{quote}

For this argument, game development is regarded as a process by which world knowledge of humans (how objects rest and collide, how spaces connect, what makes a level traversable) is written into executable digital worlds. The built assets are useful, but the development trajectory is worth more: it records what the creator intended, what the state should be, what changed, whether engine checks passed afterward, and whether the human accepted the result. This claim is also falsifiable: if training on development traces and human-engine rewards does not improve OOD generalization, then game development cannot be treated as the missing data engine for world models.

\paragraph{Contributions.} 
The primary contribution of this paper is an argument and a research agenda supported by controlled studies. First, we identify the \emph{verifiability bottleneck} and argue that the slow progress of world model post-training stems from the absence of cheap and reliable reward (\S\ref{sec:02-verifiability-thesis}--\S\ref{sec:03-crisis}). Second, we define AWoMo as a developer-centered agentic world model and propose game development as its practical recursive data engine, where executable world specifications and development traces jointly provide low-noise supervision (\S\ref{sec:04-substrate}; Appendix~\ref{sec:05-unification}). Third, the experimental section first states a validation roadmap (\S\ref{sec:06-roadmap}) and delivers experimental findings. Experimental results show that full human-engine feedback obtains the strongest result on \emph{UnitySceneBench}, pretraining on source data helps target adaptation in distribution shift, cross-engine transfer also gives positive results on Unreal and Godot~\citep{epicgames2026unreal,godot2026docs}, and AWoMo-generated environment data improves reported embodied metrics on R2R~\citep{anderson2018r2r}, Gymnasium MuJoCo~\citep{towers2024gymnasium,todorov2012mujoco}, and D4RL Gym-MuJoCo~\citep{fu2020d4rl,todorov2012mujoco} (\S\ref{sec:07-experiment}--\S\ref{sec:08-counterarguments}).

\begin{figure}[t]
\centering
\includegraphics[width=0.98\linewidth]{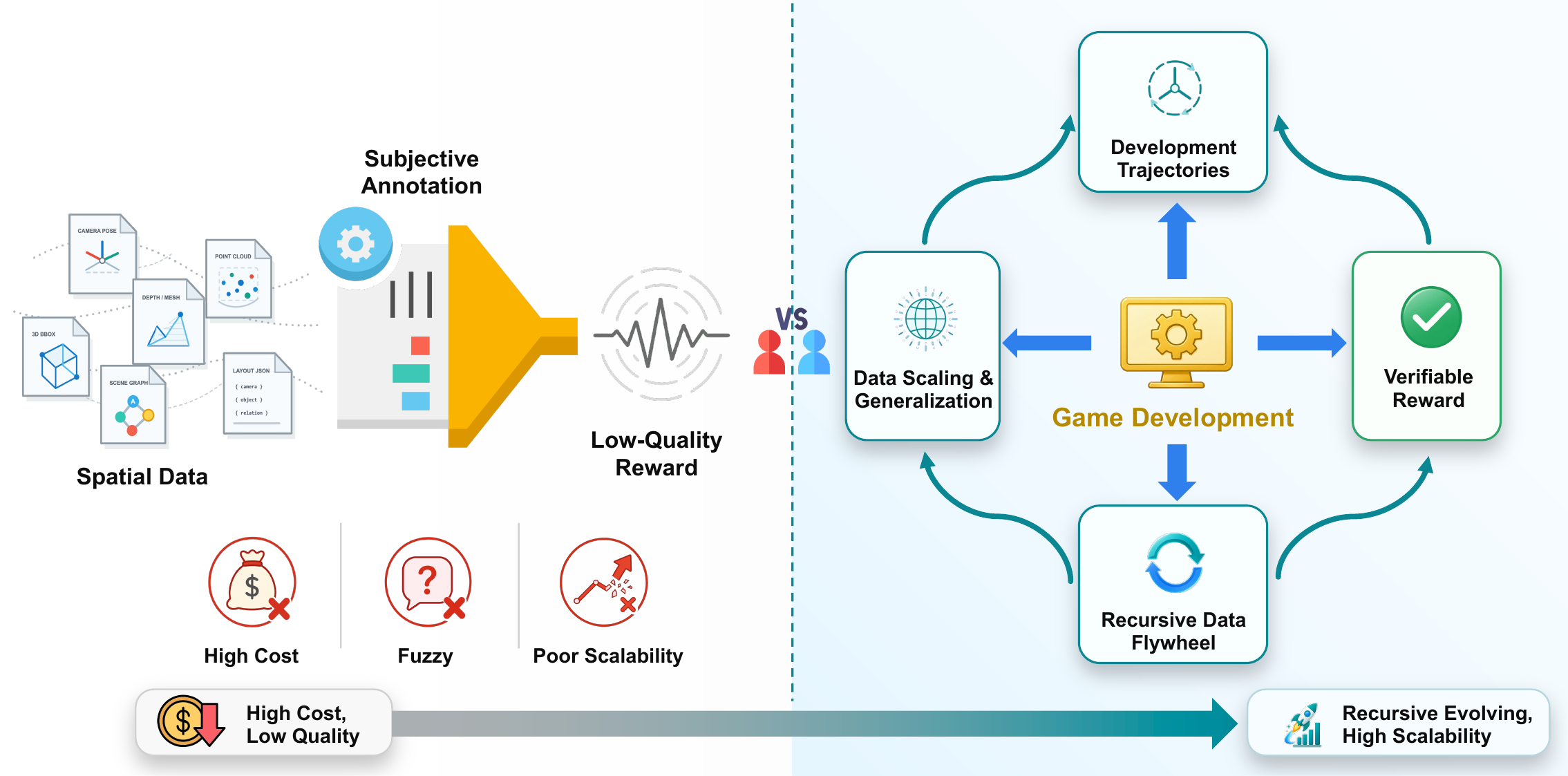}
\caption{From subjective reward proxies to a human-engine verification loop. Conventional spatial-data pipelines convert sparse human or model annotations into noisy final-output rewards. A game-development loop can instead pair human intent and feedback with engine execution, development trajectories, rendered evidence, and localized verifier failures, allowing future builders to improve from both grounded checks and developer judgment based on traces rather than final artifacts alone.}
\label{fig:verifiable-game-engine-loop}
\end{figure}

\providecommand{\RLHEV}{RLHEV}
\section{The Thesis of Verification}
\label{sec:02-verifiability-thesis}

We first separate automatic verification from full task correctness. Engines can provide the former cheaply. The latter still belongs to human intent and acceptance.

\paragraph{Verifiers and verifiable reward.}
Let $\mathcal{X}$ be the input space and $\mathcal{Y}$ the output space. A \emph{verifier} is an automatic evaluation function

\begin{equation}
V:\mathcal{X}\times\mathcal{Y}\rightarrow\mathcal{R},
\end{equation}
where $\mathcal{R}\subseteq\mathbb{R}$ denotes the reward space. Given an input $x$ and a candidate output $y$, the verifier returns a reward $r=V(x,y)$ according to an explicit correctness criterion, such as program execution, theorem checking, or numerical comparison. A task is \emph{automatically verifiable} if such a verifier can be evaluated at substantially lower cost than generating the output.

Efficient verifiers are \emph{grounded}, \emph{reliable}, and \emph{robust}: they can derive rewards from explicit task specifications rather than rely on subjective preference proxies. Reinforcement learning from verifiable reward (RLVR) uses such verifiers to train reasoning models~\citep{deepseek2025r1,lambert2024tulu3,lightman2023letsverify}. 

\paragraph{Human-engine verification.}
Previous spatial generation rarely has a complete automatic verifier. A game engine can report collider intersections, navmesh connectivity, script errors, or whether a bounded playtest probe reaches a specified goal. However, it cannot decide by itself whether a generated cutscene has the right mood. We therefore define \RLHEV{} as a new post-training framework. In the \RLHEV{}, engine checks and test suites produce dense structural labels; human acceptance verifies the true qualification. Both signals are stored in a unified protocol of state-action-check-review trajectories for post-training.

The reward abstraction of \RLHEV{} is a constrained selection over candidate outputs:

\begin{equation}
\max_{y\in\mathcal{Y}}\;\; U_H(x,y,h)-\sum_{i=1}^{n} \lambda_i(h)\,\phi_i\!\left(C_i(x,y)\right)
\quad\mathrm{s.t.}\quad
G_j(x,y)=1,\;\; j=1,\dots,m,
\end{equation}
where $x\in\mathcal{X}$ is the task input, $y\in\mathcal{Y}$ is a candidate output such as a scene program or 3D asset, and $h\in\mathcal{H}$ is the human-review context. $U_H(x,y,h)$ is the human-review utility score, typically 0 or 1 for acceptance or rejection. $C_i(x,y)\in\mathbb{R}$ are engine diagnostic checks (collision penetration, etc.), $\phi_i\ge 0$ is a monotone penalty that vanishes when check $i$ is clean, and $\lambda_i(h)\ge 0$ weights check $i$ under context $h$. $G_j(x,y)\in\{0,1\}$ ($j=1,\dots,m$) are required hard gates from the engine runtime, where $1$ denotes a pass. The invariant is the authority split: the engine offers dense, reproducible verification, while human review supplies the final judgment.

\paragraph{The recursive feedback loop.}
A verifier enables a closed optimization loop: the model proposes, feedback produces reward, the reward updates the model, and the improved model generates stronger candidates. Figure~\ref{fig:verifiable-game-engine-loop} summarizes the shift from fuzzy final-output rewards to this human-engine loop. The key property is that expensive human judgment can be grounded in cheap executable evidence rendered by a game engine, so humans can focus on final acceptance rather than on every collision or script failure. This mirrors professional game testing, where white-box tests, automated scripts, static checks, telemetry, and engine tooling coexist with black-box QA and human exploration. Self-play systems in Go and chess~\citep{silver2016alphago,silver2018alphazero} show how automatic feedback can compound improvement. As gameplay agents mature, this human-engine verification loop may further evolve into a more autonomous cycle, where one agent constructs virtual worlds while another explores, tests, and evaluates playability through gameplay, producing scalable self-improvement with less human intervention~\cite{hu2024survey,li2026meeplelm,ouyang2026gameworld}.

\paragraph{Why unverifiable signal caps data efficiency.} 

Suppose the available reward $\Rf$ is a fuzzy proxy for ground-truth quality $\Qstar$. Its error can be decomposed as
\begin{equation}
\Rf(x,y)-\Qstar(x,y)=\varepsilon(x,y)+b(x,y),
\end{equation}
where $\varepsilon$ denotes zero-mean noise and $b$ denotes systematic bias. Reward noise weakens the expected gain from each sample and lowers training efficiency at scale.
Bias is more damaging: if $\Rf$ differs from $\Qstar$ along an exploitable direction, optimizing $\Rf$ can increase proxy reward while decreasing true quality~\citep{gao2023overoptimization,skalse2022rewardhacking}. Scaling compute with such a proxy can therefore amplify the exploit. When the reward is poorly grounded, for learned capabilities the limiting factor is not only data or compute, but the fidelity and authority of the feedback signal. Engine checks improve fidelity for structural properties; human judgments preserve authority for usefulness.

\paragraph{Re-reading the Bitter Lesson.} 
The Bitter Lesson~\cite{sutton2019bitter} is often summarized as the claim that general methods which scale with compute eventually dominate hand-designed priors. This reading is incomplete for domains such as games, code, and mathematical reasoning~\cite{venkatkrishna2026aletheia,wang2026survey,stojanovski2026reasoning}. 
In these settings, scalable learning is paired with evaluators that define success: game rules, program execution, and numerical checkers. The lesson is therefore that the most efficient gain often comes from the feedback channel, instead of the model architecture design itself. Once an efficient and scalable reward mechanism exists, computation can search for candidates that satisfy it. When such reliable feedback is absent, scaling still improves imitation and perceptual plausibility, but it lacks a robust mechanism for ensuring correctness. This is the central obstacle for spatial intelligence.

\paragraph{The thesis.} 
Progress in a domain is hindered less by the lack of data or compute than by whether the domain has a scalable feedback channel. For closed tasks, this can be a cheap grounded verifier. For open-ended spatial creation, it must be human-engine verification: executable checks for what can be formalized, human judgment for the remain. Domains with strong feedback channels (games, code, mathematics) saw cost-effective, compute-scalable progress. Domains without one (video generation, 3D synthesis, world modeling) saw imitation-bound progress that is judged largely by subjective annotations over final outputs. The question for spatial intelligence is therefore not merely where to get more data, but how to instrument a workflow where efficient executable verification can be found. Section~\ref{sec:04-substrate} gives one answer: \textbf{we already build such verifiers that can help scale world models, and call them game engines.}

\section{Why Current Spatial Data Fails to Support Scaling}
\label{sec:03-crisis}

By the criterion of \S\ref{sec:02-verifiability-thesis}, progress in current spatial world models is constrained by the lack of efficient verifiers for physical and geometric correctness. We illustrate this point with three examples.

\paragraph{Video generation: impressive, but weakly grounded.} Modern video models produce realistic clips~\citep{brooks2024sora}, and interactive variants now generate playable frames in real time~\citep{bruce2024genie,deepmind2025genie3}. Yet no efficient, reliable verifier decides whether a generated video is physically and geometrically correct. \textit{Is the physics consistent? Do objects persist under occlusion? Is the perspective coherent?} In practice, video generation quality is largely judged by Fr\'echet Video Distance~\citep{unterthiner2018fvd,huang2024vbench} and human raters. These are fuzzy proxies rather than objective correctness feedback, and they can be costly, noisy, and gameable~\citep{gao2023overoptimization}.
With only fuzzy proxies over final outputs, post-training cannot be augmented with deterministic reward, leaving next-frame prediction over scraped videos as the dominant training signal.
A model can therefore learn rendering statistics without learning the hidden world state that would make 3D editing and interaction reliable~\cite{wiedemer2025video}.
It has no efficient mechanism for deterministically verifying structural failures in its own output, so gains are paid for with more data and compute~\cite{cui2026lol,liu2026improving}.
Scaling data with distribution-based objectives alone still leaves common failures such as detail misalignment, violated physical constraints, or incoherent long-horizon dynamics unsolved.

\paragraph{3D generation: little data, no cheap label.} Text-to-3D and image-to-3D have advanced quickly~\citep{poole2022dreamfusion,nichol2022pointe}, and neural scene representations now render in real time~\citep{mildenhall2020nerf,kerbl20233dgs}. But 3D generation is limited at the source: the largest curated 3D asset corpus holds on the order of $10^7$ objects~\citep{deitke2023objaverse,deitke2023objaversexl}, against the $10^{9}$ to $10^{12}$ images and tokens behind 2D vision-language and language models. The larger problem is again the verifier problem. Some local geometry properties can be checked automatically, but a high-quality 3D asset or scene of \emph{reality} still needs expensive collection, annotation and manual cleanup because there is no cheap joint verifier for physical plausibility, semantics, and task usefulness. Short on both data and reward, 3D generation often distills from 2D priors, borrowing a fuzzy signal from an adjacent unverifiable domain.

\paragraph{World simulators: the annotation wall.} Learned simulators of the physical world~\citep{ha2018world,hafner2023dreamerv3,yang2024unisim} face the hardest version of the problem. Supervising a model of world simulation needs ground truth about the physical world (depth, geometry, dynamics, contact) at a density and accuracy that is very expensive to annotate. In our view, this is a central constraint on spatial intelligence: the verifiable signal for the real world is hard to scale because it is hard to produce cheaply. Any program that depends on real-world 3D ground truth inherits this limit.

\paragraph{The unverifiability tax.}
Without a reliable verification channel, the field cannot benefit from the same post-training paradigm that drives rapid progress in code and reasoning. Progress therefore depends heavily on more data, compute, scans, and human annotations, which we refer to as the \emph{unverifiability tax}. A natural response is simply to keep paying it by collecting more videos, 3D scans, and human ratings. Compared to scaling up language models, the cost is even more enormous.

We believe a more cost-effective alternative already exists in game development. Modern game engines provide efficient and grounded verifiers for structural properties of virtual worlds, including physics, collision, navigation, rendering constraints, and bounded playability. During game development, game engines and developers naturally form a dual-verification process.
The engine verifies structural correctness, while developers judge whether the scene fulfills the world design. Grounded by engine evidence, human feedback becomes more objective and decisive. Thus, by combining efficient engine verification with developer feedback, game development offers a reliable feedback channel that can support large-scale post-training and recursive self-improvement of spatial intelligence.

\begin{figure}[t]
\centering
\includegraphics[width=0.98\linewidth]{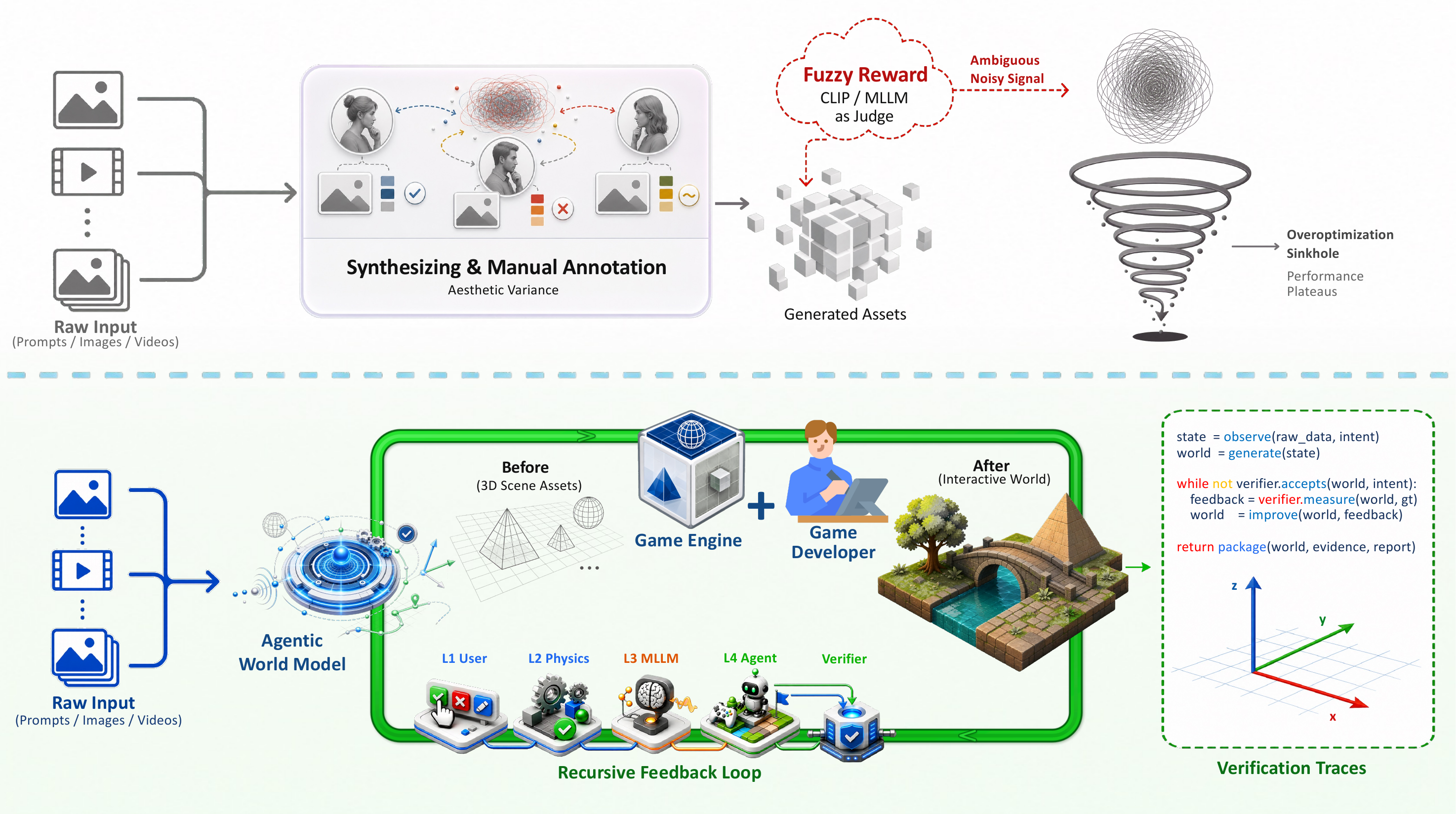}
\caption{Fuzzy reward in unverifiable spatial generation versus human-engine feedback in game development. The top pipeline illustrates how model-as-a-judge and human aesthetic variance can produce ambiguous rewards and performance bottleneck. The bottom pipeline turns assets into executable worlds: game engine and internal test suites expose local failures, rendered evidence supports review, and developer supplies the final signal for acceptance.}
\label{fig:reward-entanglement-comparison}
\end{figure}

\section{Game Development as Human-Engine Verification}
\label{sec:04-substrate}

We argue that the practical feedback already exists in game development. A scene or project encoded for a modern engine (Unity, Unreal, Godot) is an executable specification; the engine is its interpreter, runtime, and partial verifier. And a practical verifier for \emph{game development} is broader than the engine. It is the workflow in which developers define a world blueprint, inspect rendered states, run checks, playtest, critique, accept, or reject the result. Figure~\ref{fig:reward-entanglement-comparison} contrasts this human-engine loop with fuzzy final-output reward pipelines. AWoMo names this coupled system: model, agent, engine verifier, and human reviewer organized around world construction.

\paragraph{AWoMo design.}
Concretely, AWoMo is specified by four interfaces around an omni-modal world model. The \emph{intent interface} receives the task brief, references, and design constraints; the \emph{action interface} emits scene programs, asset edits, tool calls, and repair actions; the \emph{verification interface} records engine checks such as loading, collision, physics stability, navmesh reachability, script execution, and bounded playability probes; and the \emph{review interface} records developer acceptance, rejection, and critique. The execution loop is propose, render, verify, repair, and review: the model proposes an edit, the engine executes it and localizes failures, the agent issues repairs, and the loop terminates with a reviewer decision. Every loop is stored as a structured trace under the protocol introduced below, so the same workflow that builds a world also emits its own training data. Appendix~\ref{sec:A0-awomo-method} details the interfaces, the execution loop, the trace design, and the core model architecture.

\begin{figure}[t]
\centering
\includegraphics[width=0.98\linewidth]{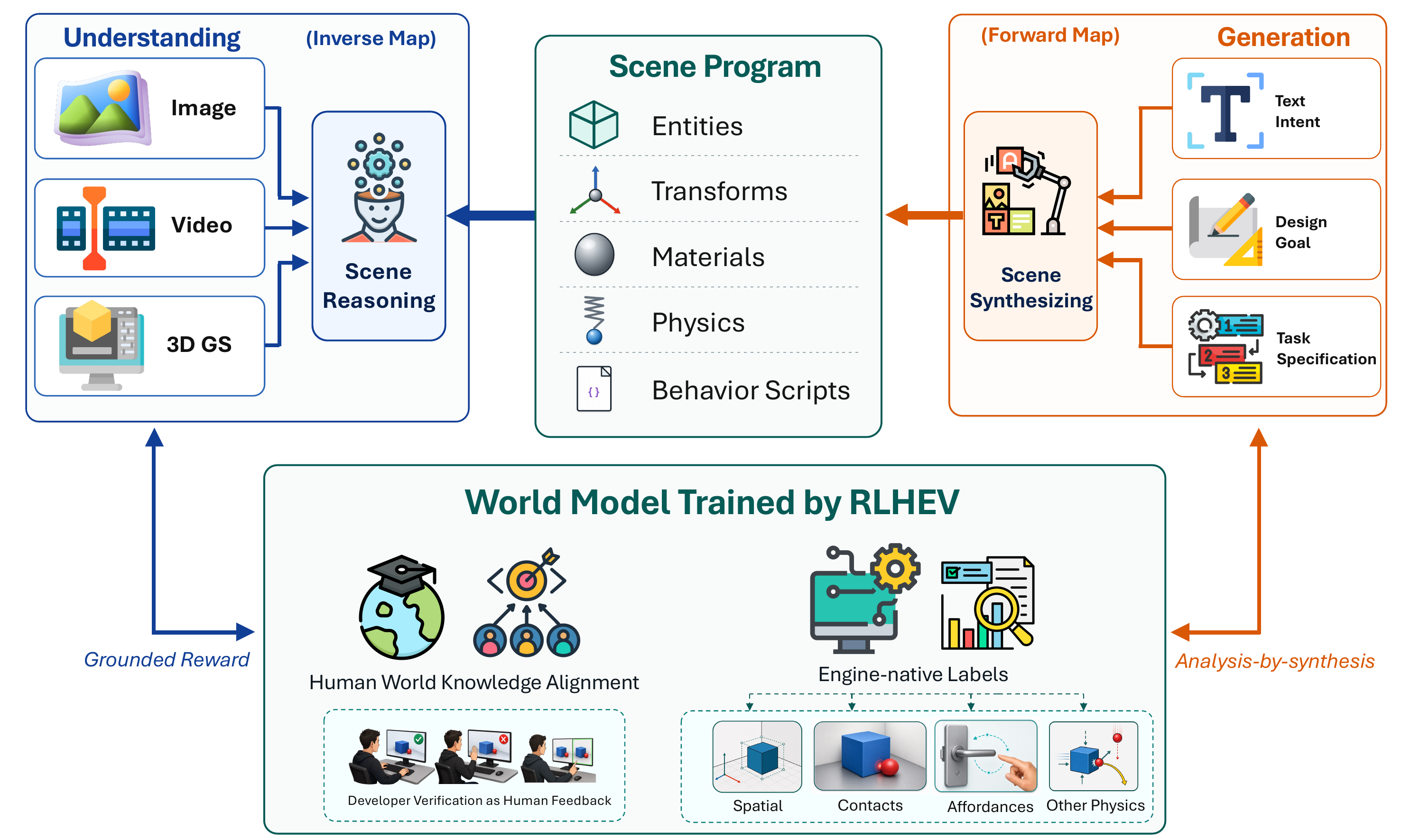}
\caption{A shared executable scene-program representation for understanding and generation. Game creation projects human world understanding, design intent, constraints, and playable experience into the model's understanding and generation capacity. Generation maps text intent, design goals, or task specifications to the scene program; understanding reads the same object in reverse. The engine checks it through rendering, physics rollouts, and temporal consistency, while humans judge whether the resulting world preserves intent and workflow value.}
\label{fig:executable-scene-program}
\end{figure}

\paragraph{What an engine checks cheaply.} As Figure~\ref{fig:executable-scene-program} shows, engine checks are grounded, lower-variance signals that separate a verifier from a proxy: geometry and collision; physics and stability; navmesh reachability; script execution and soft-locks; and bounded reachability or objective-completion probes under specified agents and seeds. They are cheap relative to manual annotation and harder to game than appearance-based metrics: a collision query is not fooled by a plausible texture.

\paragraph{Game development as a rewardable mapping.} \emph{Why would a game verifier matter for spatial intelligence?} Because game development encodes shared world knowledge of humans (how things stack, how rooms connect, what is legible) into worlds precise enough to execute.
If instrumented as the training data, games, editors, and modding ecosystems can therefore emit a difficulty-graded stream of verifiable spatial tasks whose checks inherit design intent while remaining scalable.

\paragraph{Unified World-Development Protocol (UWDP).}
A finished asset says what worked; a development trace says why it worked. Most game-development data are hard to reuse for model training because editor saves, assets, screenshots, logs, QA notes, and human decisions live in separate formats. They also lose the reason an edit was accepted or rejected. We propose \textsc{UWDP}: a typed multimodal protocol that turns ordinary game-development work into state-action-check-review traces spanning text briefs, scene programs, rendered evidence, logs, and reviewer decisions.

A compact UWDP instance is
\begin{equation}
u_t=(b,o_t,s_t,a_t,g_t,v_t,h_t,\rho_t),
\end{equation}
where $b$ is the prompt-based design intent; $o_t$ is a stable object, relation, or scene-region identifier; $s_t$ stores spatial, semantic, physical, and evidence-status fields; $a_t$ is the edit or tool action; $g_t$ is the engine or harness output; $v_t$ is rendered evidence; $h_t$ is the reviewer decision or critique; and $\rho_t$ links repairs, confidence, costs, and residual risks. Table~\ref{tab:uwdp-event-example} gives an example.

\paragraph{UWDP collection algorithm.}
The protocol is simple: (1) parse the brief and references into typed objects, relations, evidence tags, and uncertainties; (2) build a proxy or asset edit in the engine; (3) record the engine snapshot, rendered evidence, and harness checks; (4) convert failures into typed repair actions; (5) iterate until the engine test pass and reviewer return \emph{accept}; and (6) store the full trace with supervision. Figure~\ref{fig:uwdp-protocol-trace} illustrates a trace recorded by our UWDP. The resulting event can supervise next-edit prediction, preference modeling, and RL state-action-reward updates.

\paragraph{Human acceptance protocol.}
Human review remains the final supervision. UWDP separates engine-checkable fields from human-only fields: brief alignment, visual legibility, and production fit. Human rejection or acceptance is kept as an escalated review and separate labels. This keeps engine feedback dense while preventing the engine from becoming a biased oracle.

\paragraph{The analogy to code.}
A game scene is a spatial artifact that an engine makes checkable, much as compiler and tests make code checkable. The engine is the compiler and automated playtest is the test suite.
Code agents are not trusted by passing tests alone; they are trusted when delivering results align with the user's intent and review. Game development should be treated the same way. The engine signal is partial, but grounded enough to provide dense reward while keeping reward hacking measurable (\S\ref{sec:07-experiment}); human review keeps the optimization anchored to actual usefulness.

\paragraph{A ladder of reward.}
The loop can begin with cheap checks and climb toward richer ones: \emph{validity} (loads, manifold), \emph{physical plausibility} (stable rollouts), \emph{functional correctness} (navigable, objective achievable), and \emph{playability} (agent and human players can successfully complete the game with target behavior). Procedural generation and automated playtesting already instantiate parts of this ladder~\citep{summerville2018pcgml,risi2020pcg,cobbe2020procgen,makoviychuk2021isaacgym}.
Unlike the static CLIP score, this framework can keep adding bounded engine probes and human criteria as the model improves.

\paragraph{The cost asymmetry.}
In the real world, a unit of verifiable 3D signal is among the most expensive data in machine learning.
In the engine it is cheap by comparison, because the same computation that runs a world can grade it.
Human feedback remains costly, but it becomes more efficient: a developer reviews rendered candidates after automatic filters. This inverts the economics of \S\ref{sec:03-crisis}: labels are expected to become byproducts of regular development, inspection, and game testing.

\section{Experiments on Human-Engine Verification and Generalization}
\label{sec:07-experiment}

This section combines the validation roadmap with controlled experiments. The roadmap states the falsifiable claims first; the following subsections report pilot results of human-engine verification, generalization, and embodied diagnostic runs that test those claims.

\subsection{Validation Roadmap and Falsifiers}
\label{sec:06-roadmap}

To evaluate the proposed approach,
we first design oracles with measurable explicit falsifiers. We then report completed controlled studies of \textit{UnitySceneBench}, generalization, and embodied diagnostic runs, while providing findings for future studies.

\begin{figure}[t]
\centering
\makebox[0.85\linewidth][l]{%
\includegraphics[
    width=0.75\linewidth,
    trim={-2pt 0 0 0},
    clip
]{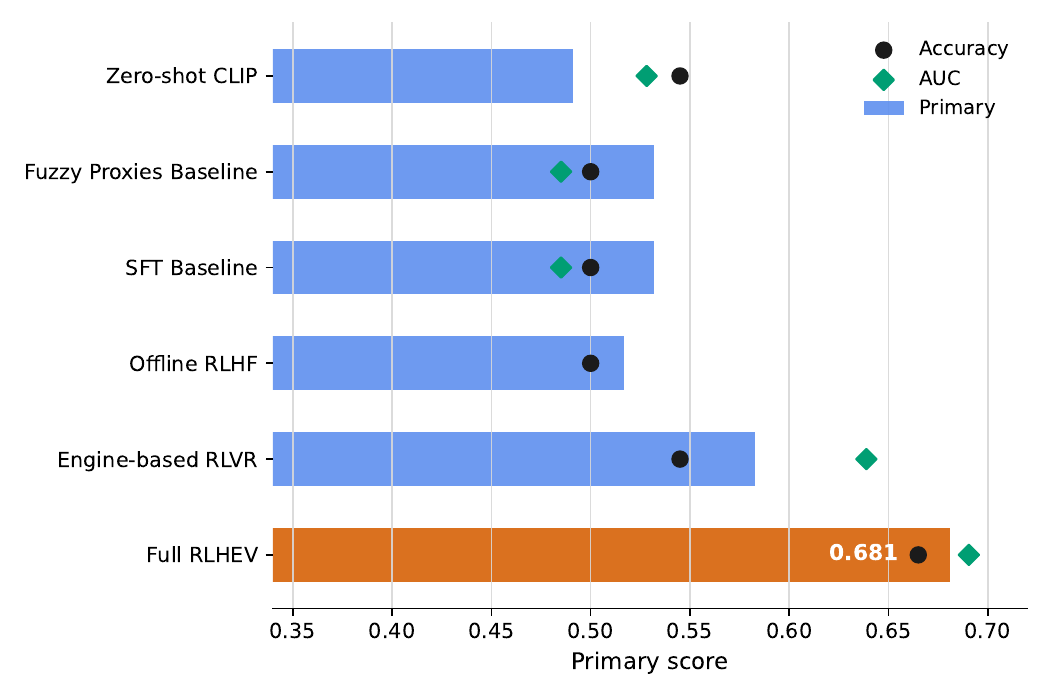}\hspace{12pt}%
}
\caption{Unity asset classification evaluation on \emph{UnitySceneBench}. Bars represent the primary score, and markers denote accuracy and AUC. This figure reports the best performance for each method across eight random seeds. Full \RLHEV{} obtains the highest best-of-eight primary score ($0.681$), with $0.665$ accuracy, $0.665$ balanced accuracy, $0.733$ F1, and $0.690$ AUC.}
\label{fig:real-rlhev}
\end{figure}

\paragraph{Oracles.}
\begin{description}[leftmargin=0em,itemsep=2pt,topsep=2pt]
\item[\textnormal{\textbf{P1 (Structural efficiency).}}] At matched compute, reinforcement learning against engine-verifiable structural rewards gives better quality for validity, physics, and playability constraints than training against fuzzy proxies (CLIP, MLLM-as-judge scores, prior preference). \emph{Falsifier:} fuzzy-reward training matches or beats verifiable-reward training at equal compute.
\item[\textnormal{\textbf{P2 (Improvement from human-engine).}}] A generator can improve through game development trajectories of propose, render, verify, human-review, repair, and retrain, as engine checks local failures and humans judge global acceptance. \emph{Falsifier:} adding human acceptance to engine traces does not improve task usefulness, sample efficiency, or robustness over engine-only traces.
\item[\textnormal{\textbf{P3 (Generalization).}}] Capabilities learned against human-engine verified trajectories can generalize when the source and target share executable structure and compatible human goals. The transfer learning test is designed as: pretrain on source world, evaluate first on held-out within-family OOD shifts and then on harder shifts in different engines. \emph{Falsifier:} no positive OOD transfer gain against controlled baselines after pre-training on source data.
\end{description}

\subsection{Experimental Setup}

We evaluate AWoMo in a human-engine workflow: the world model generates or edits game assets inside an agentic harness; engine checkers and a human reviewer evaluate the outputs; and the resulting signals are used for post-training. For brevity, detailed experiment settings are described in Appendix~\ref{sec:appendix-experiments}.

\paragraph{\emph{UnitySceneBench}.}
The multimodal understanding experiment evaluates Unity asset classification for edited asset candidates~\citep{unitydocs2026} (Appendix~\ref{sec:A1}). The multimodal inputs include the text edit prompt, Unity asset/context features, reference-image features, and structured layout payloads. We refer to our benchmark as \emph{UnitySceneBench}. The primary metric is
\begin{equation}
\text{Primary}=0.45\,\text{balanced accuracy}+0.25\,\text{accuracy}+0.20\,\text{F1}+0.10\,\text{AUC}.
\end{equation}

\paragraph{Baselines and rewards.}
Zero-shot CLIP~\citep{radford2021clip} is a strict CLIP-score classifier. The other main baselines are Fuzzy Proxies Baseline, SFT Baseline, Offline RLHF, and Engine-based RLVR (Appendix~\ref{sec:A1}). Human feedback is represented by accept/reject labels from the human-review channel (Appendix~\ref{sec:A14}); engine feedback comes from Unity-derived checks. The full model uses only human+engine reward, $0.65\,\text{human}+0.35\,\text{engine}$.

\subsection{Evaluating Understanding and Generation on UnitySceneBench}

\paragraph{Understanding-side evaluation.}
Figure~\ref{fig:real-rlhev} reports the best observed asset classification performance for each method across eight random seeds, where learned methods use 720 raw training instances. Zero-shot CLIP for classification is shown as a fixed reference. Under this best-of-eight view, Full \RLHEV{} reaches the highest asset classification performance, beating the strongest non-full baseline by $+0.098$ primary score and $+0.120$ accuracy/balanced accuracy. Appendix~\ref{sec:A1} reports the exact method definitions, evaluation protocol, and scaling details.

\begin{figure}[t]
\centering
\includegraphics[width=\linewidth]{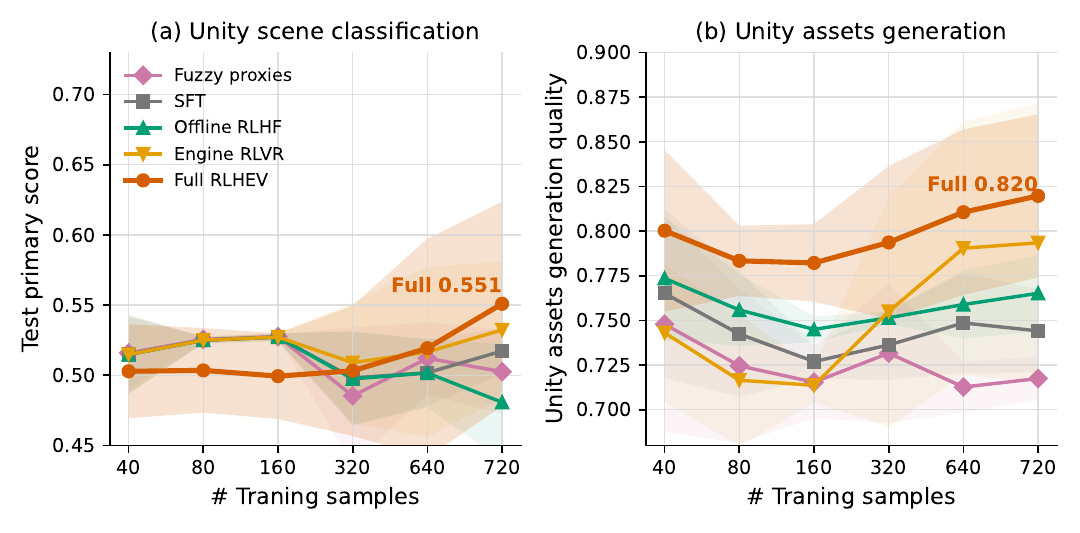}
\caption{Scaling with different numbers of training samples and tested on \emph{UnitySceneBench}.
(a) reports the primary score for Fuzzy Proxies Baseline, SFT Baseline, Offline RLHF, Engine-based RLVR, and Full \RLHEV{} on the test set, as mean$\pm$std over eight seeds. (b) Unity assets generation uses the same training budgets and reports mean$\pm$std over eight seeds per method and budget.}
\label{fig:exp1-scaling-ab}
\end{figure}

\paragraph{Scaling and generation-side evaluation.}
We next vary the amount of \emph{UnitySceneBench} training data and use the same test splits. The base run evaluates our RLHEV and baselines across different training budgets and eight random seeds per method, with training budgets up to 720 instances. Figure~\ref{fig:real-rlhev} shows the best full-budget run for each method, while Figure~\ref{fig:exp1-scaling-ab}(a) shows the mean and variance across all eight runs. As shown in Figure~\ref{fig:exp1-scaling-ab}, the generation-side task tests the same scaling question under a generation manner. In the range up to 640 training instances, Full \RLHEV{} reaches $0.8106$ generation quality; with the full 720-instance training budget, Full \RLHEV{} reaches $0.8197$ generation quality versus $0.7934$ for Engine-based RLVR.

\subsection{OOD Generalization and Cross-Engine Transfer}

\begin{figure}[t]
\centering
\includegraphics[width=\linewidth]{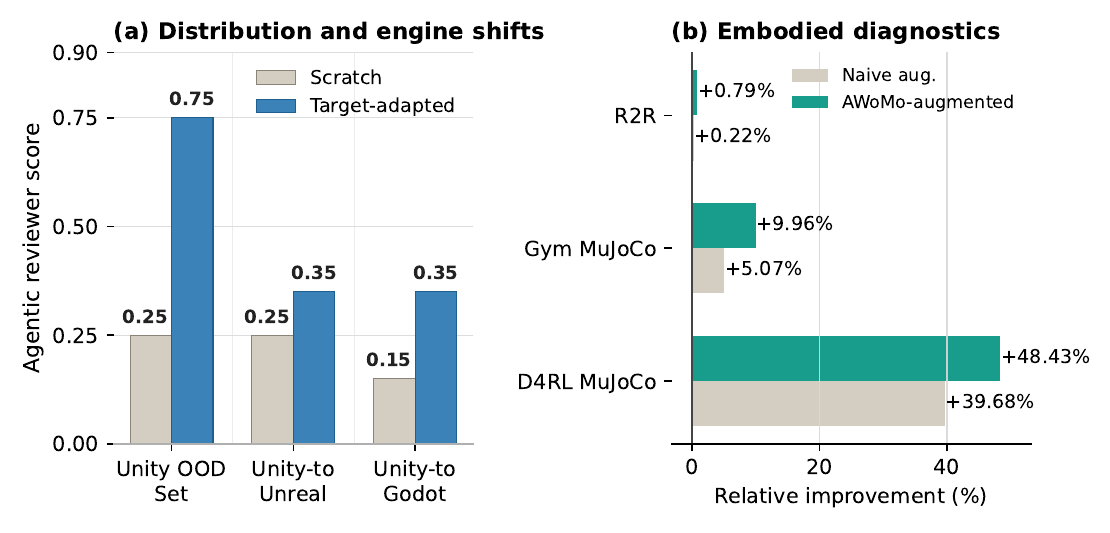}
\caption{Generalization and embodied diagnostic evaluations. (a) Generalization under distribution and engine shifts. Bars report the MLLM-as-a-judge score in $[0,1]$ for training from scratch and target-adapted from a pretrained checkpoint. Pretrained on source data and adapted on the target distribution achieve significant gain; the cross-engine settings show smaller but measurable positive signal. (b) Embodied diagnostic evaluation on R2R, Gymnasium MuJoCo, and D4RL Gym-MuJoCo. Values are relative improvements over the original baseline after normalizing scores (positive is better for all benchmarks). Naive aug. denotes standard data augmentation, whereas AWoMo-augmented denotes the policy performance trained with AWoMo-generated data.}
\label{fig:real-generalization-embodied}
\end{figure}

\paragraph{Generalization evaluation.}
We next test whether the workflow transfers across data distributions and engines (Appendices~\ref{sec:A3} and~\ref{sec:A13}). The benchmark covers Unity distribution transfer~\citep{unitydocs2026} and Unity-to-Unreal/Godot cross-engine transfer~\citep{epicgames2026unreal,godot2026docs}. The main score is the normalized MLLM-as-a-judge score in $[0,1]$, supplemented by proxy and engine metrics. We note that cross-engine asset quality has no engine-native scalar that is directly comparable across Unity, Unreal, and Godot outputs, so a rubric-based judge is currently the only practical way to quantify relative quality on a common scale. The judge is therefore used in a restricted role: it follows the same accept/reject rubric as the human-review channel, every reported judge score is reviewed and verified by human inspection (Appendix~\ref{sec:A3}), and it serves as an audited evaluation instrument rather than a training reward, which is the failure mode criticized in \S\ref{sec:intro}. We compare two conditions: target-only scratch trains only on the target-domain training split, while target-adapted transfer starts from a source-trained checkpoint and then adapts on the target-domain training split. Figure~\ref{fig:real-generalization-embodied}(a) reports the comparison.

Pretrained on the source improves the Unity distribution shift from $0.25$ to $0.75$ relative to training from scratch on the target distribution. Cross-engine transfer gives a smaller but still positive signal: Unity-to-Unreal rises from $0.25$ to $0.35$ and Unity-to-Godot from $0.15$ to $0.35$ relative to scratch, with lower loss and higher proxy scores.

\subsection{Embodied Generalization Diagnostics}

\paragraph{Embodied generalization.}
We also test whether AWoMo improves
the performance of embodied policy trained through generative environment-data augmentation (Appendix~\ref{sec:A6}). The finalized benchmark covers R2R~\citep{anderson2018r2r}, Gymnasium MuJoCo~\citep{towers2024gymnasium,todorov2012mujoco}, and D4RL Gym-MuJoCo~\citep{fu2020d4rl,todorov2012mujoco}. It compares the original baseline, a naive augmentation baseline, and AWoMo-augmented training. Figure~\ref{fig:real-generalization-embodied}(b) reports direction-normalized improvement over the original baseline.

AWoMo-augmented training improves over the original baseline on all three reported metrics: $+0.79\%$ on R2R success rate, $+9.96\%$ on Gymnasium MuJoCo rollout return, and $+48.43\%$ on D4RL Gym-MuJoCo normalized score. This supports agentic environment-data generation as an effective method for improving policy in diagnostic embodied settings.

\section{Discussion}
\label{sec:08-counterarguments}

These pilot studies support a focused claim: human-engine verification is a practical feedback source for training agentic world models. On \emph{UnitySceneBench}, combining human acceptance with engine checks gives the best asset classification performance in the best-of-eight full-budget breakdown, reaching a $0.681$ primary score with $0.665$ accuracy, while the scaling curve reports seed-averaged robustness separately. As for generalization, pretrained on source data raises the performance on OOD distributions from $0.25$ to $0.75$, with positive cross-engine gains on Unreal ($0.25$ to $0.35$) and Godot ($0.15$ to $0.35$). In embodied diagnostics, AWoMo-augmented training improves the reported metrics by $+0.79\%$ on R2R, $+9.96\%$ on Gymnasium MuJoCo, and $+48.43\%$ on D4RL Gym-MuJoCo.

The strongest gains appear where the policy is trained on human-engine dual verification; current experiment results are positive but still diagnostic, requiring larger scaling studies to verify generalization capability. The next step is to connect playable artifacts, held-out target-engine checks, held-out human acceptance, and game-agent playtesting into a tighter loop; recursive self-improvement becomes possible. Appendix~\ref{sec:A8-limitation-discussion} gives the detailed limitation analysis and future path.

\section{Related Work}
\label{sec:09-related-work}

\paragraph{Neural game engines and interactive video world models.}
Genie, GameNGen, and iVideoGPT show that learned models can produce action-conditioned interactive rollouts from video-like data~\citep{bruce2024genie,valevski2024gamengen,wu2024ivideogpt}. Newer systems such as GameFactory and GameGen-X extend this direction toward open-world and controllable game-video generation~\citep{yu2025gamefactory,che2024gamegenx}. This is the strongest contrasting route to ours~\cite{seedance2026seedance,alibaba2026happyoyster}. We view these systems as important interactive world models: they can respond to actions, but they typically do not expose the explicit state, scripts, collision semantics, or localized failure traces needed for recursive verification.

\paragraph{Structured and executable 3D generation.}
A closely related line of work represents 3D worlds as executable programs rather than opaque media artifacts. WorldCoder-Bench evaluates browser-native 3D worlds through executed Three.js programs and hidden runtime-state contracts~\citep{lu2026worldcoderbench}. VoxelCodeBench, 3DCodeBench, P3D-Bench, Code2Worlds, and PhysForge further examine code-based, parametric, dynamic, or physics-grounded 3D generation~\citep{zheng2026voxelcodebench,gao2026threedcodebench,yang2026p3dbench,zhang2026code2worlds,yang2026physforge}. These works support our central distinction: visual plausibility is not the same as executable correctness. However, we argue that to generalize to more realistic scenes, 3D-native understanding and generation capabilities still need to be improved, which motivates the game-development data engine proposed in this paper.

\paragraph{Game development as an agentic verifier and data source.}
Another related line treats natural-language game specifications and real development workflows as executable environments for agents. WorldCoder learns code world models through interaction and uses them for planning and transfer~\citep{tang2024worldcoder}, and agentic world modeling frames world models around agent decisions, environment dynamics, and self-revision under evidence~\citep{chu2026agenticworldmodeling}. GameDevBench studies real development tasks involving code changes, multimodal assets, shaders, animation, and gameplay logic~\citep{chi2026gamedevbench}. GameGen-Verifier decomposes natural-language game specifications into keypoints, injects runtime states, and checks bounded behavior assertions~\citep{jia2026gamegenverifier}. Agent2World uses adaptive multi-agent testing to generate executable symbolic world models and keeps repair trajectories for training~\citep{hu2025agent2world}. Our paper differs by making the game development itself the proposed data engine: the target object is not only a passed game or a score, but the object-linked edit/check/fix trace that can train the next builder.

\paragraph{Games as verifiable supervision and open-ended curricula.}
Games have long served as platforms for evaluating and training agents and embodied models, from ALE, Unity ML-Agents, and MineDojo to LLM/VLM-based systems such as Voyager and BALROG~\citep{bellemare2013ale,juliani2018unity,fan2022minedojo,wang2023voyager,paglieri2024balrog}. More recent work uses game rules or executable transitions as verifiable training signals: Game-RL uses game code and rules to synthesize multimodal verifiable data for VLM reasoning~\citep{tong2025gamerl}, while Game Code World Model distillation and RLVR-World use executable rules or transition checks as rewards~\citep{serapio2026gamecwm,wu2025rlvrworld}. Related curriculum methods such as UED/PAIRED and POET evolve environments to expose agent weaknesses and improve generalization~\citep{dennis2020ued,wang2019poet}. We share this recursive-data intuition, but shift the focus from playing in worlds to building them: scene programs, verifier outputs, repairs, and accepted or rejected edits become the supervision for scalable world-model training.

\section{Conclusion}
\label{sec:10-conclusion}

Spatial intelligence needs scalable feedback, not just more data. Game development provides a practical source of that feedback for supporting the post-training of \RLHEV{}: human intent and acceptance combined with engine checks, rendered evidence, and edit/check/review/fix traces. This points to an agentic world model that learns inside the workflow that builds, tests, repairs, and accepts worlds. The next future-work milestone is recursive self-improvement: agentic world models build executable worlds, game agents test and stress them, and the playtest results become training signal for the next generation of stronger world models. This is why game development matters for scaling world models: it provides an executable feedback loop in which world construction, verification, and learning can reinforce each other.

{\small
\bibliographystyle{bibstyle}
\bibliography{references}
}

\appendix
\section{Supplementary Method Details}
\subsection{Agentic World Model (AWoMo)}
\label{sec:A0-awomo-method}

\paragraph{Definition.}
AWoMo denotes our Agentic World Model system. It is not a standalone generator. It is a world model embedded in an agent workflow for building executable worlds. The system contains a model, an agent controller, a game-engine verifier, a reviewer, and a trace store. The model proposes world edits or assets; the agent executes them in the development environment; the engine checks structural constraints; the reviewer judges task fit and acceptance; the resulting assets are finalized game scenes, and the development trace can be recorded as training data.

\paragraph{System boundary.}
The general AWoMo framework is defined by four interfaces. The \emph{intent interface} receives a multimodal task brief, references, and design constraints. The \emph{action interface} emits scene programs, asset edits, tool calls, or repair actions. The \emph{verification interface} records engine checks such as loading, collision, physics stability, navmesh reachability, and bounded playability probes. The \emph{review interface} records reviewer acceptance, rejection, critique, and residual-risk notes. Thus, this boundary is important: AWoMo is the coupled workflow, not a single neural network.

\paragraph{Core model architecture and pretraining.}
The neural core inside the evaluated AWoMo instances is \texttt{UnifiedGameAssetModel}, a post-trained version of the latest Mixture-of-Transformers (MoT) model Cosmos 3~\cite{agarwal2026cosmos}. It is an omni-modal model built around modality-typed token streams, with game-asset heads for text specifications, rendered images, 3D Gaussian assets, meshes, and Unity, Godot, Unreal, and MuJoCo representations. The released checkpoint has $2.890$B parameters, hidden width $3584$, $8$ transformer layers, $4$ attention heads, and $4$ gated experts; the gating routes each typed token stream through the expert mixture, so a single backbone serves both understanding-side reading and generation-side writing of scene programs. The model is initialized from Cosmos~3~\citep{agarwal2026cosmos} and obtained through on-policy distillation (OPD) followed by continued pretraining on the AWoMo manifest. The continued-pretraining run uses $8$ A100 workers, global batch size $4096$, gradient accumulation $2$, and a production manifest with $79{,}130$ training samples, $4{,}333$ validation samples, and $4{,}282$ test samples; the best checkpoint is selected at step $589$ by validation loss. Baseline and evaluation details are given in Appendix~\ref{sec:A14}.

\paragraph{Pretraining data format and authorization.}
The AWoMo pretraining manifest contains $87{,}745$ accepted samples and $504$ rejected samples. Each entry is a UWDP-style multimodal record with a task brief, one or more typed inputs, typed outputs, optional engine tokens, and metadata for source, split, and sampling weight. Output modalities include text ($87{,}745$ instances), images ($53{,}426$), Unreal ($9{,}678$), Unity ($7{,}423$), 3D assets ($3{,}194$), meshes ($3{,}194$), MuJoCo ($2{,}946$), Godot ($2{,}685$), and 3D Gaussian assets ($12$). The developer-workflow data are collected through UWDP in an Agentic Game Development product. All developer-workflow data used in the experiments are developer-authorized; accepted manifest instances are kept only after the local allowlist and quality filters pass.

\paragraph{Execution loop.}
The basic loop is propose, render, verify, repair, and review. Given intent $b$, the agent produces an edit $a_t$ and an executable state $s_t$. The engine returns check outputs $g_t$, rendered evidence $v_t$, and failure localization when checks fail. The agent can then issue a repair action, such as moving an object, replacing an asset, changing a collider, regenerating a layout region, or revising a script. The loop stops when the reviewer accepts, rejects, or requests further revision. In the intended full system, gameplay agents can also run bounded playtests before final review.

\paragraph{Trace design.}
AWoMo stores each loop as a UWDP trace rather than a final snapshot. A trace contains the design intent, stable object identifiers, intermediate scene states, actions, verifier outputs, rendered evidence, human decisions, repair links, costs, and residual risks. Stable identifiers let the model connect a failed check to the object or region that caused it and to the repair that fixed it. This makes the data closer to a code trajectory with tests and patches than to a static image-caption pair.

\begin{table}[t]
\centering
\small
\setlength{\tabcolsep}{3pt}
\caption{Example fields in one UWDP instance. The values illustrate how stable object IDs connect intent, scene state, engine checks, repairs, rendered evidence, and review.}
\label{tab:uwdp-event-example}
\begin{tabularx}{\linewidth}{@{}p{0.10\linewidth}p{0.36\linewidth}X@{}}
\toprule
Field & Example value & Functionality \\
\midrule
$b$ & Arrange a playable storage room with a clear path to the exit. & Conditions generation on design intent. \\
$o_t$ & \texttt{crate\_07}; relation \texttt{near(chair\_03)}. & Localizes checks and repairs to objects or regions. \\
$s_t$ & Transform, collider bounds, material tags, navmesh status, and visibility flags. & Grounds the model in executable scene state. \\
$a_t$ & Move \texttt{crate\_07} to $(2.4,0.0,2.6)$ and regenerate its collider. & Supervises next-edit and repair prediction. \\
$g_t$ & Collision fail before repair; navmesh pass after repair. & Supplies verifier gates and dense rewards. \\
$v_t$ & Rendered preview, object mask, and camera metadata. & Provides reviewer-visible evidence. \\
$h_t$ & Reviewer marks \emph{needs revision}, then \emph{accept} after the repair. & Provides acceptance and preference signal. \\
$\rho_t$ & Link from failed collision to repair action; edit cost; residual risk note. & Supports credit assignment and risk-aware selection. \\
\bottomrule
\end{tabularx}
\end{table}

\paragraph{From traces to training data.}
Each stored trace is compiled into training instances in three forms. Accepted terminal states become supervised generation targets conditioned on the intent $b$; failed checks paired with their repair actions become next-edit and repair-prediction pairs; and the engine outputs $g_t$ together with the reviewer decision $h_t$ become the fused reward of \RLHEV{}, with engine gates applied before the human-engine combination. Rejected traces are kept as well: they supply the negative half of the acceptance signal that final-artifact corpora usually lose.

\paragraph{Training signal.}
RLHEV is the training objective used for AWoMo when human and engine feedback are available. Engine checks supply dense local rewards or gates; human review supplies the acceptance target and veto. The same trace can also train supervised next-edit prediction, next-failure prediction, repair-success prediction, preference models, and candidate rankers.

\paragraph{RLHEV objective.}
Let $\mathbf{x}$ denote the multimodal state and task context, $\mathbf{a}$ is a generated edit or action sequence, $h(\mathbf{x},\mathbf{a})\in[0,1]$ the human-review reward, and $e(\mathbf{x},\mathbf{a})\in[0,1]$ the normalized engine reward. Let $\mathbf{g}(\mathbf{x},\mathbf{a})\in\{0,1\}^m$ be binary engine gates and $\mathbf{1}$ the all-one vector. The gated scalar reward used by Full-\RLHEV{} is
\begin{equation}
r_{\alpha,\beta}(\mathbf{x},\mathbf{a})
=
\mathbb{I}\{\mathbf{g}(\mathbf{x},\mathbf{a})=\mathbf{1}\}
\left(\alpha h(\mathbf{x},\mathbf{a})+\beta e(\mathbf{x},\mathbf{a})\right)
-\lambda_c c(\mathbf{x},\mathbf{a}),
\end{equation}
where $c(\mathbf{x},\mathbf{a})\ge 0$ is an optional cost or risk penalty. In the UnitySceneBench main result, $\alpha=0.65$, $\beta=0.35$, and $\lambda_c=0$ unless a candidate is rejected by a required gate. The Offline RLHF ablation sets $(\alpha,\beta)=(1,0)$; Engine-based RLVR sets $(0,1)$.

Let $\pi_{\theta}(\mathbf{a}\mid\mathbf{x})$ be the trained policy and $\pi_0$ the supervised reference policy. A standard regularized offline objective is
\begin{equation}
\max_{\theta}\;
\mathbb{E}_{(\mathbf{x},\mathbf{a})\sim\mathcal{D}}
\left[
w(\mathbf{x},\mathbf{a})
\left(r_{\alpha,\beta}(\mathbf{x},\mathbf{a})-b(\mathbf{x})\right)
\log \pi_{\theta}(\mathbf{a}\mid\mathbf{x})
\right]
-\lambda_{\mathrm{KL}}
\mathbb{E}_{\mathbf{x}\sim\mathcal{D}}
\left[
D_{\mathrm{KL}}\!\left(\pi_{\theta}(\cdot\mid\mathbf{x})\,\|\,\pi_0(\cdot\mid\mathbf{x})\right)
\right],
\end{equation}
where $w(\mathbf{x},\mathbf{a})\ge 0$ is an offline importance or selection weight, $b(\mathbf{x})$ is a scalar baseline, and $\lambda_{\mathrm{KL}}\ge 0$ keeps the policy near the supervised model. The reported Unity experiment implements this principle through offline reward-weighted candidate training and held-out selection.

\paragraph{Remark on reward-estimation error.}
A standard uniform-error argument applies to the implemented reward: if the implemented reward deviates from the intended reward by at most $\epsilon$ everywhere, then the value gap of an $\eta$-optimal policy under the implemented reward is at most $2\epsilon+\eta$ under the intended reward. 
% We record this only as a reminder that reducing reviewer and engine-label error directly tightens the worst-case policy-selection gap; it is not a claim about full game quality or open-domain human preference alignment.

\subsection{Unifying Understanding and Generation of Scene Assets}
\label{sec:05-unification}

As shown in Figure~\ref{fig:executable-scene-program}, our human-engine protocol is to place generation and understanding around the same scene-program representation. This perspective aligns with the broader trend toward unified models that perform multimodal understanding and generation within one framework~\citep{lu2022unifiedio,team2024chameleon,zhou2024transfusion,chen2025januspro}.

\paragraph{A scene is a program; the engine is its interpreter.} Represent a 3D world not as pixels or an opaque latent but as a structured, executable specification: entities, transforms, materials, physical properties, and behavior scripts. The separated spatial understanding and generation tasks can therefore become inverse directions of one map:
\begin{itemize}[leftmargin=1.4em,itemsep=1pt,topsep=2pt]
\item \textbf{Generation} is the forward map from design or synthesis intent to scene program or assets: produce a world that satisfies a goal.
\item \textbf{Understanding} is the inverse map from observation of scene program, image, video, or scan to the reasoning results of the world scenes.
\end{itemize}

The two directions also share their supervision. The action that creates a generation target also creates much of its understanding supervision. The development protocol also adds the temporal and human dimensions: an observation can be paired not only with a final scene program, but with the edits, verifier failures, rendered previews, critiques, and accept/reject decisions that led to it. Passive real-world data rarely offers this density of causal labels; here it is a long-overlooked byproduct of running and reviewing the program.

\paragraph{Why this helps both directions.} Understanding and generation are expected to complement each other when both operate on the same executable scene program~\cite{wu2025janus,zhou2025opening}. A model that generates stable, navigable scenes must learn the spatial structure needed to understand them; a model that recovers program intents from observations gains the prior needed to know how generated scenes behave correctly. This mirrors inverse-graphics systems, where an encoder recovers a structured scene representation and a decoder or renderer uses that representation for generation and manipulation~\citep{yao20183daware}.

\paragraph{Closing the world-model loop.} 
The engine makes structural constraints cheap and repeatable to test, while human review decides whether the resulting world actually satisfies the intended purpose. The goal is therefore not to claim that game-engine success automatically transfers to all spatial domains. Rather, the claim is that executable feedback and human judgment together provide a scalable training loop for world models. Whether this loop generalizes beyond games still should be tested through further controlled studies.

\subsection{Protocol Trace Analysis}
\label{sec:A7}

\paragraph{UWDP Trace Illustration.}
\label{sec:A7-uwdp-trace}

\begin{figure}[t]
\centering
\includegraphics[width=\linewidth]{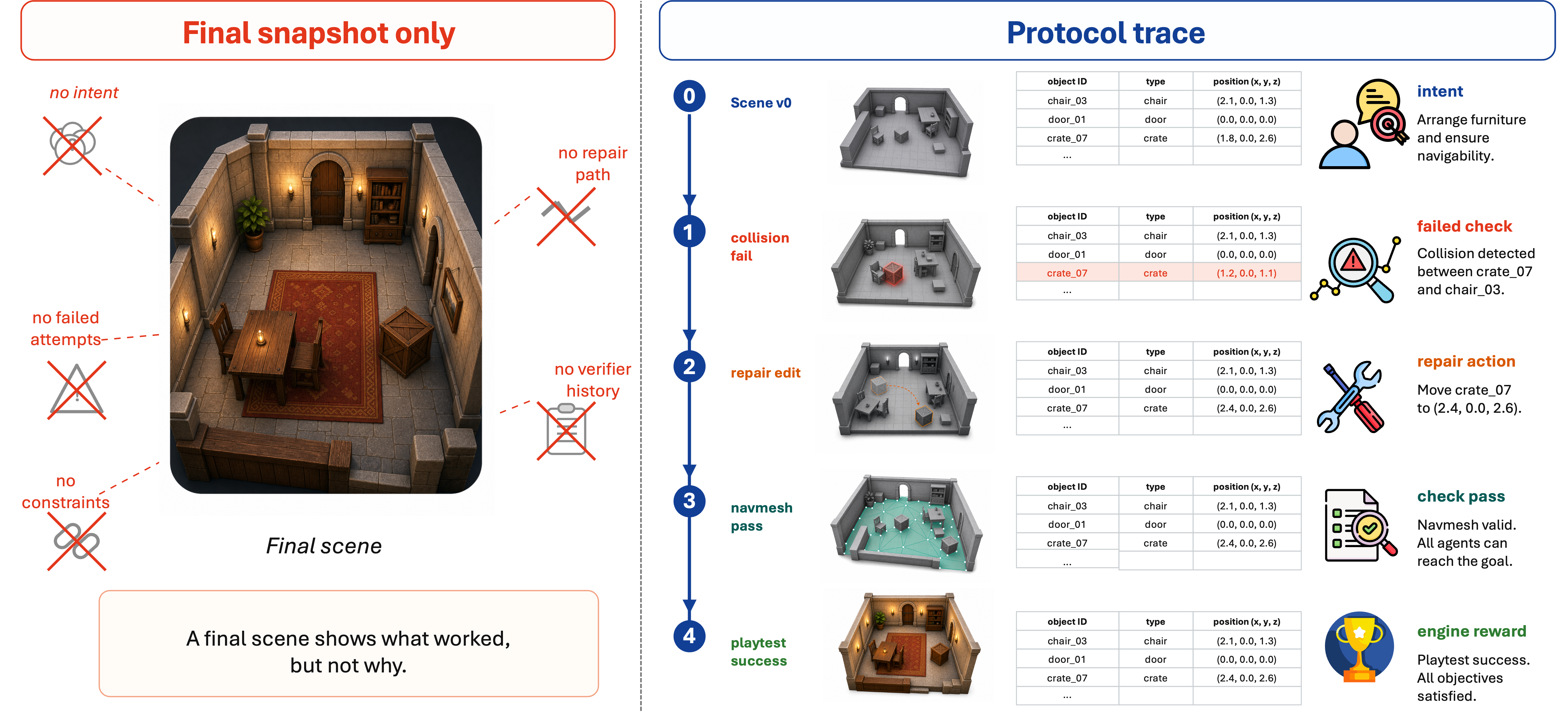}
\caption{Protocol traces preserve the process data that is absent from final snapshots. A final scene records only the terminal state, whereas a trace records intent, intermediate scene states, failed checks, repair edits, validation outcomes, and engine rewards. This object-linked sequence provides a training objective for asset diagnosis and repair.}
\label{fig:uwdp-protocol-trace}
\end{figure}

Figure~\ref{fig:uwdp-protocol-trace} illustrates why a final scene is not enough supervision. A snapshot shows what worked, but not intent, failed attempts, constraints, verifier history, or repair paths.

\paragraph{Source-Data Scaling.}
\label{sec:A7-source-scaling}

This analysis tests whether source-side workflow traces help target-engine ranking under a small target-label budget. The source split contains 720 Unity training examples, 80 validation, and 200 test examples~\citep{unitydocs2026}. The Godot and Unreal target splits~\citep{godot2026docs,epicgames2026unreal} each contain 720 training, 80 validation, and 200 test examples with target-engine labels. For each seed and target engine, the probe is trained under a deliberately scarce target-label budget: among the target-engine training examples, only 8 randomly sampled instances expose their target-engine pass/warn labels to the probe, and all remaining target instances are treated as unlabeled. This simulates the practical setting where a new engine has just been instrumented and almost no verified target-engine labels exist yet, so the probe must rely on transferred source-side signal. Under this budget, we vary the number of Unity source examples from 0 to 720.

\paragraph{Representations.}
The \emph{snapshot-only} probe uses coarse final-output metadata such as format tags, prompt length, asset count, and aggregate output size. It does not use labels, deviation type, asset identity, or target-engine checks. The \emph{protocol-trace} probe adds source-side workflow fields: Unity accept/reject label, deviation type, target-engine id, and used-asset identity. It does not use hidden target-engine check values or target-engine reason strings. Both probes are ridge-regression ranking models evaluated by Spearman correlation against held-out target-engine pass/warn labels.

\begin{figure}[t]
\centering
\includegraphics[width=\linewidth]{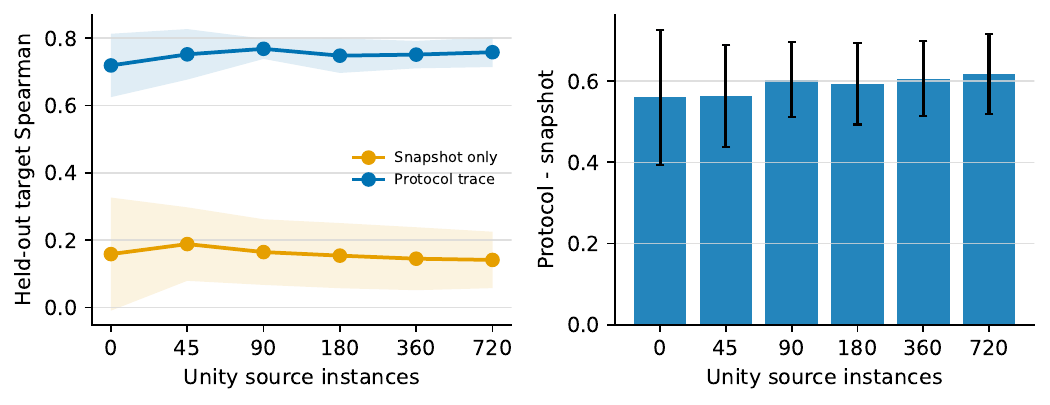}
\caption{Source-side trace transfer under an 8-label target budget. Values are held-out Spearman correlations over 8 seeds and two target engines. Protocol-trace features provide substantially stronger target-engine ranking signal than coarse final-snapshot features; increasing the number of Unity source instances gives a smaller additional gain.}
\label{fig:real-source-scaling}
\end{figure}

\paragraph{Result.}
Figure~\ref{fig:real-source-scaling} shows a clear representation gap. With development traces, snapshot-only features reach $0.159\pm0.168$ Spearman, while protocol traces reach $0.719\pm0.094$. Using 720 source instances, snapshot-only remains low ($0.141\pm0.084$), while protocol traces reach $0.758\pm0.044$. The trace advantage stays around $0.56$--$0.62$ across source sizes.

\section{Supplementary Experiment Details}
\label{sec:appendix-experiments}
\subsection{Baselines, Benchmarks, and Evaluation Details}
\label{sec:A14}

This appendix collects baseline, benchmark, and evaluation details for the reported experiments. It is intended to keep the main text concise while making the experimental claims auditable. Model architecture, pretraining, and \RLHEV{} objective details are given in Appendix~\ref{sec:A0-awomo-method}.

\paragraph{Reviewer components and baselines.}
The zero-shot visual baseline uses CLIP~\citep{radford2021clip}. The agentic MLLM-as-a-judge harness uses Qwen3.6-35B-A3B~\citep{qwen36_35b_a3b} under the same accept/reject rubric used by the human-review channel; its scores are used for evaluation and audit only, and all released judge scores were reviewed and verified by human inspection. Engine checks, held-out labels, and test examples remain separate from model inputs.

\paragraph{UnitySceneBench data.}
UnitySceneBench instances are constructed from the developer workflow (Appendix~\ref{sec:A0-awomo-method}) at a smaller evaluation scale. A task prompt and asset context are converted into a candidate edit, imported into Unity, rendered, checked by the engine harness, and assigned a reviewer accept/reject label.

\paragraph{Benchmarks and engines.}
Unity, Unreal, and Godot are cited through their engine documentation~\citep{unitydocs2026,epicgames2026unreal,godot2026docs}. The embodied diagnostics use R2R~\citep{anderson2018r2r}, Gymnasium MuJoCo~\citep{towers2024gymnasium,todorov2012mujoco}, D4RL Gym-MuJoCo~\citep{fu2020d4rl,todorov2012mujoco}, and SAME/DUET-style VLN trajectory evaluation~\citep{zhou2025same,chen2022duet}.

\subsection{Full-\RLHEV{} Validation Details}
\label{sec:A1}

This appendix describes the full-\RLHEV{} evaluation on \emph{UnitySceneBench}.

\paragraph{Task and data.}
The task is binary asset classification for edited Unity asset candidates~\citep{unitydocs2026}: the model predicts whether a candidate edit is accepted or rejected. The input includes edit prompts, Unity asset/context features, reference-image features, candidate-layout features, and optional CLIP features~\citep{radford2021clip}. The target label and engine verdict are excluded from the model input. The split has 720 training examples, 80 validation examples, and a 200-example test set with 100 accepted and 100 rejected candidates.

\paragraph{Rewards and baselines.}
Zero-shot CLIP~\citep{radford2021clip} is used as a fuzzy proxy. Human feedback is represented by accept/reject labels from the human-review channel, while engine feedback comes from Unity-derived checks. The full-\RLHEV{} setting combines only human and engine reward with weights $0.65$ and $0.35$, respectively. Appendix~\ref{sec:A14} describes the reviewer-channel convention. Table~\ref{tab:rlhev-method-list} gives the definition of baseline methods.

\begin{table}[t]
\centering
\scriptsize
\setlength{\tabcolsep}{3pt}
\caption{Method definitions for the \emph{UnitySceneBench} comparison.}
\label{tab:rlhev-method-list}
\begin{tabular}{@{}p{0.27\linewidth}p{0.66\linewidth}@{}}
\toprule
Name & Definition \\
\midrule
Zero-shot CLIP & A visual-proxy baseline. It classifies the acceptance or rejection of candidates using a threshold on CLIP similarity, without a learned task model or RL. \\
Fuzzy Proxies Baseline & The baseline model trained with CLIP similarity rewards. It tests whether visual-proxy rewards help the supervised baseline. \\
SFT Baseline & The baseline model trained only with supervised GT accept/reject labels from the training split. The GT is the binary accepted/rejected target label. It is excluded from the model input, and no RL reward is used. \\
Offline RLHF & Offline RL using only the human-review reward, without engine reward. \\
Engine-based RLVR & Offline RL using only engine-verification reward, without the human-review reward. \\
Full \RLHEV{} & The full model trained with fused human-review and engine rewards, using weights $0.65$ and $0.35$. \\
\bottomrule
\end{tabular}
\end{table}

\paragraph{Evaluation protocol.}
For Figure~\ref{fig:real-rlhev}, the learned methods are trained with the 720 raw instances and evaluated on the same 200-instance test split under eight random seeds. The figure reports the best observed full-budget test performance for each method across these eight runs, so it should be read as a best-of-eight metric breakdown rather than a seed-averaged robustness estimate. Figure~\ref{fig:exp1-scaling-ab} reports the complementary mean$\pm$std view across training budgets and summarizes seed-level stability.

The generation-side protocol uses the same training budgets rather than a generated-candidate horizontal axis. The 720 full-RLHEV generation run produces 640 fresh Unity plans and labels with Unity exit code 0 and no missing render/snapshot/layout/engine-label artifacts. Table~\ref{tab:asset-generation-by-training} breaks down the same-protocol generation quality for Figure~\ref{fig:exp1-scaling-ab}.

\begin{table}[t]
\centering
\scriptsize
\setlength{\tabcolsep}{3.0pt}
\caption{Unity assets generation quality for Figure~\ref{fig:exp1-scaling-ab}. Values are mean quality over eight seeds per method for each training budget.}
\label{tab:asset-generation-by-training}
\begin{tabular}{@{}lcccccc@{}}
\toprule
Method & 40 & 80 & 160 & 320 & 640 & 720 \\
\midrule
Fuzzy Proxies Baseline & 0.7478 & 0.7246 & 0.7154 & 0.7316 & 0.7127 & 0.7174 \\
SFT Baseline & 0.7651 & 0.7423 & 0.7270 & 0.7362 & 0.7486 & 0.7441 \\
Offline RLHF & 0.7735 & 0.7558 & 0.7449 & 0.7515 & 0.7589 & 0.7652 \\
Engine-based RLVR & 0.7430 & 0.7165 & 0.7137 & 0.7550 & 0.7904 & 0.7934 \\
Full \RLHEV{} & \textbf{0.8002} & \textbf{0.7833} & \textbf{0.7821} & \textbf{0.7936} & \textbf{0.8106} & \textbf{0.8197} \\
\bottomrule
\end{tabular}
\end{table}

\paragraph{Claim boundary.}
This experiment evaluates a bounded asset classification verifier for Unity asset edits. In the best-of-eight full-budget breakdown, Full \RLHEV{} is the strongest method, with a $+0.098$ primary-score gain and $+0.120$ accuracy/balanced-accuracy gain over the strongest non-full baseline. The multi-seed scaling curve in Figure~\ref{fig:exp1-scaling-ab}(a) should be used for seed-averaged robustness. The result does not imply final game quality or broader human preference alignment.

\subsection{Generalization Experiment Details}
\label{sec:A3}

This appendix describes the generalization evaluation.

\paragraph{Benchmark categories.}
The benchmark is intentionally split into two forms of generalization.
\begin{itemize}[leftmargin=1.4em,itemsep=1pt,topsep=2pt]
\item \emph{Distribution Transfer}: Unity source distribution to held-out Unity examples~\citep{unitydocs2026}. The split contains 5{,}895 source examples and 1{,}254 target examples.
\item \emph{Cross-Engine Transfer}: Unity source assets to Unreal and Godot targets~\citep{unitydocs2026,epicgames2026unreal,godot2026docs}. Each source/target split contains 1{,}000 examples with 720/80/200 train/validation/test partitions.
\end{itemize}

\paragraph{Training conditions.}
Target-only scratch uses only the target-domain training split and does not initialize from a source-domain checkpoint. Target-adapted transfer first trains on the source-domain split, then adapts that checkpoint on the target-domain training split. All reported scores are measured on the target-domain test split. When reported, zero-shot evaluates the base checkpoint directly, and source-only evaluates the source-trained checkpoint without target adaptation. For the cross-engine boosted runs, target-adapted transfer starts from the Unity source checkpoint and runs 160 target-adaptation steps on the target-engine training split.

\paragraph{Metrics.}
The main metric is a normalized MLLM-as-a-judge score in $[0,1]$. The MLLM-as-a-judge harness judges whether the generated asset is usable, engine-ready, and aligned with the task. There is currently no engine-native scalar metric that is directly comparable across Unity, Unreal, and Godot outputs, so this rubric-based judge is the only practical common scale for relative quality across engines. To keep the judge trustworthy, it follows the same accept/reject rubric as the human-review channel, and every reported judge score was reviewed and verified by human inspection before release; the judge is used for evaluation only and never as a training reward. We also report proxy scores and loss as independent checks alongside the MLLM-as-a-judge score. Table~\ref{tab:generalization-full} reports the aggregate values. For cross-engine transfer, the target entries report target-adapted transfer.

\begin{table}[t]
\centering
\scriptsize
\setlength{\tabcolsep}{2.5pt}
\caption{Generalization aggregate. Target means target-adapted transfer; scratch means target-only scratch training.}
\label{tab:generalization-full}
\begin{tabular}{@{}llccccc@{}}
\toprule
Category & Transfer & Condition & Assert & Proxy & Engine acc. & Loss \\
\midrule
Distribution & Unity $\rightarrow$ held-out Unity & zero-shot & $0.25$ & $0.451$ & $0.007$ & $11.685$ \\
Distribution & Unity $\rightarrow$ held-out Unity & target & $\mathbf{0.75}$ & $0.449$ & $0.008$ & $11.590$ \\
Distribution & Unity $\rightarrow$ held-out Unity & scratch & $0.25$ & $0.421$ & $0.005$ & $11.909$ \\
Cross-engine & Unity $\rightarrow$ Unreal & target & $0.35$ & $0.474$ & $0.008$ & $11.596$ \\
Cross-engine & Unity $\rightarrow$ Unreal & scratch & $0.25$ & $0.437$ & $0.009$ & $11.825$ \\
Cross-engine & Unity $\rightarrow$ Godot & target & $0.35$ & $0.478$ & $0.007$ & $11.575$ \\
Cross-engine & Unity $\rightarrow$ Godot & scratch & $0.15$ & $0.441$ & $0.008$ & $11.820$ \\
\bottomrule
\end{tabular}
\end{table}

\paragraph{Interpretation.}
Distribution transfer is the strongest positive case. Target adaptation raises the MLLM-as-a-judge score from $0.25$ to $0.75$ on the Unity distribution shift.

Cross-engine transfer also shows positive signal. Pretraining on the source and applying target-adapted transfer improves Unity-to-Unreal from $0.25$ to $0.35$ and Unity-to-Godot from $0.15$ to $0.35$ relative to scratch. Proxy scores and losses also improve. We use these results as generalization evidence in a transfer learning manner: source-engine traces provide useful initialization, and target-engine adaptation turns them into improved target-engine signal.

\subsection{Cross-Engine Generalization}
\label{sec:A13}

The cross-engine evaluation tests whether a Unity-trained asset workflow transfers to Unreal or Godot. Unity~\citep{unitydocs2026}, Godot~\citep{godot2026docs}, and Unreal~\citep{epicgames2026unreal} are evaluated with their corresponding target-engine labels.

\paragraph{Setup.}
The source and target splits each contain 1{,}000 instances with a 720/80/200 train/validation/test split. Multimodal inputs include text, image, Unity, and mesh modalities. Outputs are text plus the target engine representation. The conditions follow Appendix~\ref{sec:A3}: scratch uses only the target-engine training split, while target-adapted transfer starts from the Unity source checkpoint and adapts on the target-engine training split. Table~\ref{tab:cross-engine-real} reports the target-adapted transfer result for each target engine.

\begin{table}[t]
\centering
\scriptsize
\setlength{\tabcolsep}{3pt}
\caption{Cross-engine target-adapted transfer results. Target-adapted transfer gives measurable positive signal over scratch for both target engines, with higher proxy scores and lower losses.}
\label{tab:cross-engine-real}
\begin{tabular}{@{}llccc@{}}
\toprule
Transfer & Condition & MLLM judge & Proxy & Loss \\
\midrule
Unity $\rightarrow$ Unreal & zero-shot & $0.25$ & $0.396$ & $12.377$ \\
Unity $\rightarrow$ Unreal & Unity-source only & $0.15$ & $0.005$ & $16.693$ \\
Unity $\rightarrow$ Unreal & target-adapted transfer & $0.35$ & $0.474$ & $11.596$ \\
Unity $\rightarrow$ Unreal & scratch & $0.25$ & $0.437$ & $11.825$ \\
Unity $\rightarrow$ Godot & zero-shot & $0.25$ & $0.396$ & $12.375$ \\
Unity $\rightarrow$ Godot & Unity-source only & $0.15$ & $0.005$ & $16.695$ \\
Unity $\rightarrow$ Godot & target-adapted transfer & $0.35$ & $0.478$ & $11.575$ \\
Unity $\rightarrow$ Godot & scratch & $0.15$ & $0.441$ & $11.820$ \\
\bottomrule
\end{tabular}
\end{table}

\paragraph{Result.}
Target-adapted transfer raises Unity-to-Unreal from $0.25$ to $0.35$ and Unity-to-Godot from $0.15$ to $0.35$ relative to scratch. It also improves proxy score and loss for both target engines. These instances support the paper's engineering claim: an agentic world model should carry engine-specific protocol traces and calibrate to the target runtime before deployment.

\subsection{Embodied Generalization Details}
\label{sec:A6}

This appendix describes the embodied generalization evaluation. The reported scope covers R2R~\citep{anderson2018r2r}, Gymnasium MuJoCo~\citep{towers2024gymnasium,todorov2012mujoco}, and D4RL Gym-MuJoCo~\citep{fu2020d4rl,todorov2012mujoco}. Each benchmark compares three methods: the original baseline, a naive augmentation baseline, and AWoMo-augmented training.

\paragraph{Method definition.}
AWoMo denotes our proposed method of Agentic World Model. In the embodied experiment, AWoMo is not a standalone policy architecture; it is a profile-guided data-augmentation workflow driven by the released AWoMo checkpoint profile. For R2R, the workflow scores existing PREVALENT augmentation trajectories with a selector derived from the checkpoint profile, matches the resulting subset to the target path/instruction distribution with scan balancing, and fine-tunes the SAME policy on it. For Gymnasium MuJoCo and D4RL Gym-MuJoCo, the workflow synthesizes candidate state-action pairs by profile-guided augmentation and filters them with automatic acceptance checks before policy training.

\paragraph{Metrics.}
The three benchmarks use different metrics, so Figure~\ref{fig:real-generalization-embodied}(b) reports direction-normalized improvement over the original baseline. R2R uses success rate, Gymnasium MuJoCo uses rollout return, and D4RL Gym-MuJoCo uses the D4RL normalized score; higher is better for all three main metrics. Table~\ref{tab:embodied-final} reports the raw values.

\begin{table}[t]
\centering
\scriptsize
\setlength{\tabcolsep}{3pt}
\caption{Final embodied generalization results. Values are mean$\pm$std where available. $\Delta$ is the direction-normalized relative improvement of AWoMo-augmented training over the original baseline.}
\label{tab:embodied-final}
\begin{tabular}{@{}llccccc@{}}
\toprule
Benchmark & Metric & Original & Naive aug. & AWoMo-aug. & $\Delta$ & Best \\
\midrule
R2R & Success rate $\uparrow$ & $76.006{\pm}0.133$ & $76.176{\pm}0.175$ & $76.607{\pm}0.094$ & $+0.79\%$ & AWoMo \\
Gymnasium MuJoCo & Rollout return $\uparrow$ & $1568.44{\pm}1756.69$ & $1648.03{\pm}1689.15$ & $1724.73{\pm}1642.60$ & $+9.96\%$ & AWoMo \\
D4RL Gym-MuJoCo & Normalized score $\uparrow$ & $18.295{\pm}14.663$ & $25.555{\pm}12.448$ & $27.155{\pm}9.491$ & $+48.43\%$ & AWoMo \\
\bottomrule
\end{tabular}
\end{table}

\paragraph{Result.}
AWoMo-augmented training improves over the original baseline on all three reported metrics. The largest relative gain is on D4RL Gym-MuJoCo, where the normalized score improves by $+48.43\%$. Gymnasium MuJoCo rollout return improves by $+9.96\%$, and R2R success rate improves by $+0.79\%$. The R2R experiment uses SAME/DUET trajectories~\citep{zhou2025same,chen2022duet} on the val-unseen split. The D4RL experiment uses Gymnasium MuJoCo rollouts with the official D4RL normalization constants.

\section{Supplementary Discussion}
\subsection{Limitation Discussion and Future Analysis}
\label{sec:A8-limitation-discussion}

We test the proposed approach against its strongest objections.

\paragraph{C1: Games are not reality (the sim-to-real gap).} Correct; the gap is the central test, not an assumption. Simulation already transfers some policies to reality through domain randomization and large-scale training~\citep{tobin2017domainrand,makoviychuk2021isaacgym,savva2019habitat}; here the question is which executable structures survive OOD shifts. Our current experiments do not contain real scans, real robots, or a genuine real-to-sim-to-real loop, so they should not be read as evidence that game-engine rewards already transfer to the physical world. The next right test is explicit: use real-world environment, adapt with a little bit finetuning, and measure which executable checks remain useful. Until then, games are a scalable verifier-rich substrate, not a solved bridge to reality.

\paragraph{C2: Video generation is already advancing fast.} It is, and the effect is impressive~\citep{brooks2024sora,deepmind2025genie3}. But the progress is imitation- and compute-bound, judged by argument and fuzzy aggregates, with no self-improvement loop. Its typical failures are the physical and long-horizon inconsistencies. Progress on plausibility is not progress on correctness.

\paragraph{C3: The engine reward can be gamed too.} Yes. A game engine is a partial verifier, not an oracle. A loose collision, navigation, or budget check can be optimized around if it is used alone. This is not a reason to abandon verifiers. It is a reason to use the combination of verifier ensembles, randomized probes, held-out checks, and human review. Engine rewards are still useful because their failures are localizable and testable rather than only perceptual.

\paragraph{C4: Real-world 3D data will get cheap.} Cheaper data is not a cheaper verifier. A scan shows what one instance looked like without annotation. It also cannot tell whether a new synthesized output is correct. The engine supplies that missing automatic check; passive capture does not.

\paragraph{C5: Source-engine traces may overfit to one engine.}
Different engines encode different import formats, collision semantics, navigation systems, and runtime constraints, so target calibration matters. The cross-engine study (Appendix~\ref{sec:A13}) shows measurable positive signal for Unity-to-Unreal and Unity-to-Godot transfer after target adaptation, with stronger proxy scores and lower losses. The result supports the role of source-engine traces as useful initialization data and target runtime calibration as the step that turns them into target-engine signal.

\paragraph{Evidence and validation boundary.}
The experiments give bounded evidence for the roadmap. Human-engine feedback is strongest on the \emph{UnitySceneBench} asset classification and generation tasks, and target adaptation plus agentic environment data improve the reported generalization and embodied diagnostics. These results do not yet prove complete game quality, human-subject validity, or a closed-loop embodied deployment. The next validation step is to implement playable artifacts and consented human studies.

\paragraph{Future paradigm and path.}
The long-term goal is a recursive game-building loop. An agentic world model builds an executable game; game agents play it, find broken rules, unreachable goals, physics failures, and weak pacing; those failures become training signal for the next world model. The main difficulty is making the loop trustworthy: the game specification must be executable, the playtest agents must be hard to fool, and human review can still judge design intent. A practical path is to start with bounded playable levels, add playtest agents and failure localization, then scale to multi-scene games with new engines, agents, and human reviewers.

\subsection{Broader Impacts}
\label{sec:A2}

This paper advocates a research direction rather than releasing a model, so its impacts are indirect. If this approach is effective, grounding spatial generation in verifiable engine reward would lower the cost of training spatial intelligence systems and embodied AI (with benefits for environment simulation and policy training) and would shift effort from scraping large real-world corpora toward computation against automatic verifiers, with attendant data-governance benefits. The same capabilities carry the usual dual-use risks of better world and video generation, including more convincing synthetic media. An engine-grounded pipeline is, if anything, easier to audit than an opaque generative model, since its outputs are produced against explicit, inspectable checks. At the same time, real development traces may contain creator intent, proprietary assets, debugging decisions, and project IP; deployed data engines should therefore use opt-in collection, project-level filtering, redaction of irrelevant content, access controls, and local or federated training paths for sensitive projects. Finally, raising capability through verifiable reward makes verifier design more important: a misspecified engine reward would be optimized just as literally as a fuzzy one.

The approach also reframes recursive self-improvement. Much recent discussion imagines capability growth arising primarily from AI systems training, evaluating, and improving other AI systems. Our thesis points instead to a hybrid loop in which AI systems participate in human productive processes and learn from the executable traces those processes generate. In game development, human spatial knowledge is continually externalized into model capabilities; agentic systems can contribute to this process while learning from it. Recursive improvement is therefore not only an AI-to-AI phenomenon, but can also be an ecosystem property emerging from the interaction between human practice, executable environments, and learning systems. Whether such socially grounded feedback loops scale beyond game development to broader domains of engineering, science, and physical-world interaction is an open research question.

Looking further ahead, the same loop suggests a demand-side flywheel. As scaling-law-driven models keep making software and content production dramatically cheaper, a growing share of human time and ambition is freed to move into digital worlds, where people can design, build, and inhabit virtual environments that express their own ideas and dreams. Each of these creators can work through an agentic world-building pipeline of the kind described in this paper, in which intent, edits, engine checks, and acceptance decisions are recorded as reusable interactive 3D construction traces. World-building activity then compounds: cheaper creation invites more creators, more creators produce a superlinearly growing corpus of verified construction traces, and those traces in turn accelerate the scaling of the next generation of world models. In this long-term view, the data engine is not a fixed corpus but a growing economy of human creativity in virtual worlds, whose natural byproduct is exactly the verified supervision that world models need.

\end{document}